\documentclass[conference]{IEEEtran}
\IEEEoverridecommandlockouts
\usepackage{cite}
\usepackage{amsmath,amssymb,amsfonts}
\usepackage{algorithmic}
\usepackage{graphicx}
\usepackage{textcomp}
\usepackage{xcolor}
\usepackage{authblk}
\def\BibTeX{{\rm B\kern-.05em{\sc i\kern-.025em b}\kern-.08em
    T\kern-.1667em\lower.7ex\hbox{E}\kern-.125emX}}
\begin{document}

\title{\huge Revisiting Data Augmentation in Model Compression:\\
An Empirical and Comprehensive Study
}


\author{Muzhou Yu\textsuperscript{1*}\thanks{*Equal contribution}, 
Linfeng Zhang\textsuperscript{2*}, 
Kaisheng Ma\textsuperscript{2\dag}\thanks{\dag Corresponding author: kaisheng@mail.tsinghua.edu.cn}}

\affil{\textsuperscript{1} Xi'an Jiaotong University, 
\textsuperscript{2} Tsinghua University}


\maketitle

\begin{abstract}
The excellent performance of deep neural networks is usually accompanied by a large number of parameters and computations, which have limited their usage on the resource-limited edge devices. To address this issue, abundant methods such as pruning, quantization and knowledge distillation have been proposed to compress neural networks and achieved significant breakthroughs. However, most of these compression methods focus on the architecture or the training method of neural networks but ignore the influence from \emph{data augmentation}. In this paper, we revisit the usage of data augmentation in model compression and give a comprehensive study on the relation between model sizes and their optimal data augmentation policy.
   To sum up, we mainly have the following three observations:
    \textbf{(A)} Models in different sizes prefer data augmentation with different magnitudes. Hence, in iterative pruning, data augmentation with varying magnitudes leads to better performance  than data augmentation with a consistent magnitude.
    \textbf{(B)}
    Data augmentation with a high magnitude may significantly improve the performance of large models but harm the performance of small models. Fortunately, small models can still benefit from  strong data augmentations by firstly learning them with ``additional parameters'' and then discard these ``additional parameters'' during inference.
    \textbf{(C)} The prediction of a pre-trained large model can be utilized to measure the difficulty of data augmentation. Thus it can be utilized as a criterion to design better data augmentation policies.
    We hope this paper may promote more research on the usage of data augmentation in model compression.
\end{abstract}


\section{Introduction}
In the last decade, extensive breakthroughs have been accomplished with deep neural networks in various tasks and domains~\cite{vit,bert}. However, modern neural networks usually contain massive parameters which make them unaffordable in practical applications, especially on resource-limited edge devices. 
Recently, abundant model compression methods have been proposed to address this problem by means of efficient network architecture~\cite{deepcompression,mobilenetv2,shufflenet,squeezenet}, low-precision representation~\cite{actication_quantization,INQ,deepcompression} and better training method~\cite{distill_hinton, kd_ab, kd_bert1, selfdistillation,kd_crd}.
Concretely, neural network pruning iteratively removes the unimportant neurons in neural networks to search for not only accurate but also efficient architectures. Quantization methods strive to represent the weight and activation of neural networks with fewer bits. Knowledge distillation is proposed to improve the performance of small models by transferring the knowledge from a pre-trained teacher model. Abundant experimental and theoretical breakthroughs have been achieved with these methods in various tasks, such as classification~\cite{distill_hinton, deepcompression}, object detection~\cite{detectiondistillation,thundernet,kd_detection1,kd_detection2}, segmentation~\cite{structured_kd,3d_segment,yolact} and language models~\cite{kd_bert1,kd_bert2,wu2020lite}.

Unfortunately, another crucial role in deep learning - \emph{data} has usually been ignored in the study of model compression. It is generally acknowledged that the quality and diversity of training data have an essential and fundamental impact on the performance of neural networks.
However, most previous model compression research studies the compression on models in isolation from the influence of data.
In particular, their works usually apply the same data augmentation policy for models of different sizes during compression.
In this paper, to uncover the relation between \emph{data augmentation} and \emph{model compression}, an empirical and comprehensive analysis has been introduced. In summary, we mainly have the following three observations.

\textbf{\emph{Models in different sizes prefer data augmentation with different magnitudes.}} Previous automatic data augmentation policies usually search an optimal data augmentation policy for a specific dataset. However, we find that the optimal data augmentation policies for the models in different sizes are also different. Usually, a small model tends to benefit from a weak data augmentation (\emph{i.e.} data augmentation with a low magnitude) but get harmed by a strong data augmentation  (\emph{i.e.} data augmentation with a high magnitude). In contrast, a large model can earn more benefits from a strong data augmentation policy. We suggest that this is because the regularization effect from a strong data augmentation may be too overlarge to be learned for a small model. Based on this observation, we propose to gradually reduce the magnitude of data augmentation during iterative network pruning instead of using the consistent data augmentation policy for models in different pruning ratios.

\textbf{\emph{Small models can learn strong data augmentations with additional parameters.}} Although small models can not directly benefit from a strong data augmentation, we find that they can still learn knowledge from a strong data augmentation in an indirect manner with \emph{additional parameters}. 
For instance, in neural network pruning,
by being initialized with the weights of the pre-pruning model which is trained with a strong data augmentation, the pruned model can inherit the knowledge learned from a strong data augmentation and thus achieves higher performance. Similarly, it is also possible for small models to firstly learn knowledge from a strong data augmentation with some additional layers, and then drop these additional layers during inference. Interestingly, our experimental results demonstrate that although additional layers are dropped, the knowledge from the strong data augmentation learned by them can still be preserved by the small model.

\textbf{\emph{Large models can be used to find the data augmentation policies for small models.}} Motivated by the intuition that \emph{if a data augmentation is too strong to be learned for a large model, it should also be too strong for a small model}, we show that pre-trained large models can be utilized to find the optimal data augmentation for a small model. For instance, in knowledge distillation, the teacher model can be utilized to select the data augmentation, which maximizes the knowledge distillation loss to accelerate model convergence and sometimes lead to better generalization ability.

These observations have demonstrated that the usage of data augmentation is closely relevant to the capacity of neural networks. We hope this paper can promote more research to study model compression and data augmentation together.

\section{Related Work}
\subsection{Data Augmentation}
Data augmentation, which aims to improve the diversity of the training set by applying multiple predefined transformations, has become one of the essential techniques in machine learning. In computer vision, random image cropping and horizontal flipping are the two most popular augmentation methods for both supervised learning ~\cite{resnet,Alexnet} and unsupervised learning~\cite{simclr,moco}. Besides, Cutout and Random Erasing methods are proposed to randomly remove some pixels in the images~\cite{cutout,randomerasing,yang2019region}. Mixup and CutMix are proposed to synthesize new images as the linear combination of existing two images~\cite{cutout,zhang2017mixup}.  Besides common visual tasks, the effectiveness of data augmentation has also been witnessed in natural language processing~\cite{dataaug_text1,dataaug_text2,dataaug_text3}, AI fairness~\cite{dataaug_fair}, model robustness~\cite{dataaug_robust,Benchmarking_segmentation,BenchmarkingNN_classifiction}, single image super resolution~\cite{dataaug_superresolution} and so on.

Recently, motivated by the success of neural network architecture searching, abundant methods have been proposed to automatically find the optimal data augmentation policy for a given dataset. AutoAugment is firstly proposed to parameterize each image transformation with probability and magnitude parameters~\cite{AutoAugmentLA}. Then, RandAugment is proposed to reduce the searching space by using a global magnitude parameter~\cite{rand}. Besides, abundant methods have been proposed to accelerate the searching process with adversarial learning~\cite{adversarial_pruning,fasteraug}, population algorithm~\cite{ho2019population}, meta learning~\cite{dada,ddas}, and automatic hyper-parameter tuning~\cite{ohl}.
Unfortunately, most previous research in data augmentation ignores the fact that the optimal data augmentation policies for different neural networks are also different. RandAugment shows that usually the neural network with fewer layers requires data augmentation with a lower magnitude~\cite{rand}. Fu~\emph{et al.} show that in knowledge distillation, the optimal data augmentation policies for students and teachers are different~\cite{kd_aug}. Recently, Suzuki~\emph{et al.} propose TeachAugment, which aims to filter the hard data augmentation for students by employing a pre-trained teacher model~\cite{suzuki2022teachaugment}. These research motivates us to conduct a comprehensive study on how to apply data augmentation in model compression.

\vspace{-0.15cm}

\subsection{Neural Network Pruning}
The rapidly increasing performance of deep neural networks is usually accompanied by enormous amounts of computational and storage overhead. To address this issue, neural network pruning is proposed to remove redundant parameters from an existing neural network. 
In the last century, pruning has already been proposed to delete the unimportant neurons with the criterion of their Hessian matrix~\cite{optimal_brain_damage,hassibi1993second}. Recently, extensive works have been proposed to prune the filters~\cite{filter_prune}, channels~\cite{channel_prune}, patterns~\cite{niu2020patdnn}, and layers~\cite{layer_prune} in convolutional neural networks~\cite{deepcompression}, recurrent neural networks~\cite{lstm_prune1}, Transformers~\cite{vit_prune1,vit_prune2} in terms of $L_0$ distance~\cite{pruing_l0}, geometric median~\cite{geometric_prune}, meta learning~\cite{metapruing}, and reconstruction error~\cite{luo2017thinet}. Besides classification, pruning methods have also been introduced in more challenging tasks such as object detection, segmentation, pretrained language models, and image-to-image translation~\cite{segment_prune1,segment_prune2,detect_prune1,detect_prune2,detect_prune3,gan_compress,teacher_do_more_gankd,wang2020gan}

\vspace{-0.15cm}

\subsection{Knowledge Distillation}
Knowledge distillation, also known as student-teacher learning, has become one of the most effective neural network training methods for model compression and model performance boosting. It is firstly proposed by Bucilua~\emph{et al.} ~\cite{model_compression} to compress ensemble neural networks for data mining. Then, Hinton~\emph{et al.} propose the concept of knowledge distillation which aims to compress a single neural network by training a lightweight student network to mimic the prediction results (\emph{e.g.} categorical probability) of the teacher network~\cite{distill_hinton}. Since the student network inherits the knowledge from its teacher, it usually can achieve much higher performance than traditional training.
Recently, extensive following-up methods have been proposed to distill teacher in not only teacher logits, but also teacher knowledge in backbone features and the invariants, such as attention~\cite{attentiondistillation}, relation~\cite{relational_kd,relational_kd2}, positive value~\cite{kd_comprehensive}, task-oriented information~\cite{tofd} and so on. Besides classification, it has also achieved excellent performance in object detection~\cite{detectiondistillation,kd_detection3,kd_detection2,kd_detection1,kd_detection4,local_kd}, semantic segmentation~\cite{structured_kd}, image-to-image translation~\cite{gan_compress,spkd_gan,omgd,revisit_discriminator,wkd}, machine translation~\cite{lin2020weight}, pretrained language models~\cite{kd_bert1,kd_bert2}, multi-exit models~\cite{selfdistillation,Zhang2019SCANAS}, model robustness~\cite{auxiliarytraining,kd_defense} and so on.

\textbf{Data Augmentation in Knowledge Distillation} Recently, a few works have been proposed to study the usage of data augmentation in knowledge distillation. Fu~\emph{et al.} propose to apply different data augmentation policies to students and teachers to facilitate knowledge distillation~\cite{fu2020role}.
Das~\emph{et~al.} have given an empirical study on the effectiveness of data augmentation in knowledge distillation, which finds that the mix-based data augmentation methods harms the performance of students~\cite{das2020empirical}. Wang~\emph{et al.} find that knowledge distillation loss can tap into the extra information from different input views brought by data augmentation~\cite{wang2020knowledge}. Wei~\emph{et al.} propose to circumvent the outliers of Autoaugment~\cite{AutoAugmentLA} with an additional knowledge distillation loss~\cite{wei2020circumventing}.

\vspace{-0.15cm}
\subsection{Other Model Compression Techniques}
Besides neural network pruning and knowledge distillation, recently, some other model compression techniques have also been proposed, including quantization, lightweight model design, neural architecture search, and dynamic neural networks. Quantization aims to represent the weight and activation of neural networks with eight and even less bits~\cite{quantization_2,actication_quantization,data_free_quantization}. Lightweight models are proposed to reduce the complexity of neural networks by using lightweight building blocks, such as depthwise-separable convolution, group convolution, shuffle convolution and so on~\cite{mobilenetv2,mobilenets,shufflenet,addernet,song2020addersr}. Neural architecture search is proposed to search the most efficient and accurate architecture of neural networks based on reinforcement learning, meta learning, or the other differential optimization methods~\cite{Zhang2019SCANAS,slimmable,slimmablev2,metapruing}. Dynamic neural networks are proposed to inference each sample with instance-adaptive resolution, channels, or depths~\cite{NAS_kd,mobilenet3,efficientdet}.

\section{Preliminaries}
In this paper, we mainly focus on the supervised classification setting, where a training set $\mathcal{X}=\{x_1,...,x_n\}$ and the corresponding one-hot label set $\mathcal{Y}=\{y_1,...,y_n\}$ is given.
Let $c$ be the number of possible classes and $f:\mathcal{X}\times\mathbf{W}\rightarrow \mathbb{R}^c$ as a classifier parameterized by $\mathbf{W}$, then its training objective with cross entropy loss $\mathcal{L}_{\text{CE}}$ can be formulated as 
\begin{equation}
    \label{crossentropy}
    \mathop{\min}_{\mathbf{W}} \mathcal{L}_{\text{CE}}\Big(f(\mathcal{X};\mathbf{W}), \mathcal{Y} \Big) = - \mathop{\sum }_{i=1}^n\mathop{\sum}_{j=1}^c y_{i,j}\log \sigma_j \Big(f(x_i;\mathbf{W})\Big), 
\end{equation}
where $y_{i,j} =1$ if $x_i$ belongs to $j$-th category. $\sigma_j$ indicates the softmax value on $j$-th element. 

\subsection{RandAugment}
In this paper, we adopt RandAugment~\cite{rand} as the data augmentation policy for all the experiments. Given a set of $K$ atomic data augmentation operations $\{o_1, o_2, ..., o_K\}$ and a global magnitude parameter $M$, the augmentation process with RandAugment $\mathcal{A}(x, M)$ can be formulated as 
\begin{equation}
    \mathcal{A}(x; M) =  o_i\Big(o_j(x;M), M\Big),
\end{equation}
where $o_i$ and $o_j$ are randomly sampled from $\{o_1, o_2, ..., o_K\}$. $o_i(x,M)$ indicates apply $o_i$ augmentation with magnitude $M$ to $x$\footnote{In this paper, we fix the number of used data augmentation as 2 for convenience.}. During the searching phrase of RandAugment, $\mathcal{X}$ and $\mathcal{Y}$ are divided into non-overlapped training set   $\mathcal{X_{\text{train}}}$ and $\mathcal{Y_{\text{train}}}$, and the validation set $\mathcal{X}_{\text{val}}$ and $\mathcal{Y}_{\text{val}}$, respectively. Then, the searching objective for the optimal magnitude $M^*$ can be formulated as 
\begin{equation}
\label{randaug}
    \begin{aligned}
    &M^*=\mathop{\arg\min}_{M} \mathcal{L}_{\text{CE}}\Big(f(\mathcal{A}(\mathcal{X}_{\text{val}};M);\mathbf{W^*}), \mathcal{Y}_{\text{val}}\Big) \\
    &\text{s.t.}~~\mathbf{W^*}=\mathop{\arg\min}_{\mathbf{W}} \mathcal{L}_{\text{CE}}\Big(f(\mathcal{A}(\mathcal{X}_{\text{train}};M);\mathbf{W}), \mathcal{Y}_{\text{train}}\Big).
    \end{aligned}
\end{equation}
Note that although Equation~\ref{randaug} is a bi-level optimization problem, the searching space of $M$ is the integers from 0 to 30. Thus, this optimization problem can be solved even with grid searching. 

\subsection{Knowledge Distillation}
In the typical knowledge distillation proposed by Hinton~\emph{et al.}~\cite{distill_hinton}, the teacher model is usually pre-trained with the training loss described in Equation~\ref{crossentropy}. Then, a student is trained to mimic the Kullback-Leibler divergence between its outputs and the prediction of teachers. By using scripts $\mathcal{S}$ and $\mathcal{T}$ to distinguish students and teachers, the knowledge distillation loss can be formulated as
\begin{equation}
\begin{aligned}
 &\mathcal{L}_{\text{KL}}\Big(f_{\mathcal{T}}(\mathcal{X};\mathbf{W}_{\mathcal{T}}), f_{\mathcal{S}}(\mathcal{X};\mathbf{W}_{\mathcal{S}}) \Big)\\
 &= - \tau^2 \mathop{\sum }_{i=1}^n\mathop{\sum}_{j=1}^c  \sigma_j \Big(f_\mathcal{T}(\mathcal{X};\mathbf{W})/\tau\Big) \log\sigma_j\Big(f_\mathcal{S}(\mathcal{X};\mathbf{W})/\tau\Big),
\end{aligned}
\end{equation}
and the overall training loss of the student can be formulated as $\alpha \cdot \mathcal{L}_{\text{CE}} + (1-\alpha) \cdot \mathcal{L}_{\text{KL}}$, where $\alpha \in (0,1]$ is a hyper-parammeter to balance the two loss functions.  
\subsection{Neural Network Pruning}
Neural network pruning aims to reduce the number of parameters in neural networks by removing the unimportant neurons. Its training objective can be formulated as 
\begin{equation}
\begin{aligned}
    \mathop{\min}_{\mathbf{W}} \mathcal{L}_{\text{CE}}\Big(f(\mathcal{X};\mathbf{W}),\mathcal{Y}\Big)~~~~~~~~\text{s.t.}~~\frac{\text{Card}(\mathbf{W})}{\text{Num}(\mathbf{W})} < 1-p,
\end{aligned}
\end{equation}
where Card() and Num() return the number of nonzero elements and all the elements in $\mathbf{W}$, respectively. $p$ indicates the desired pruning ratio. A larger $p$ indicates that more parameters are removed from the neural network. In this paper, we adopt the built-in  $L_1$-norm unstructured pruning method in Pytorch~\cite{pytorch} for all our experiments.

\begin{figure*}
\begin{center}
    \includegraphics[width=\linewidth]{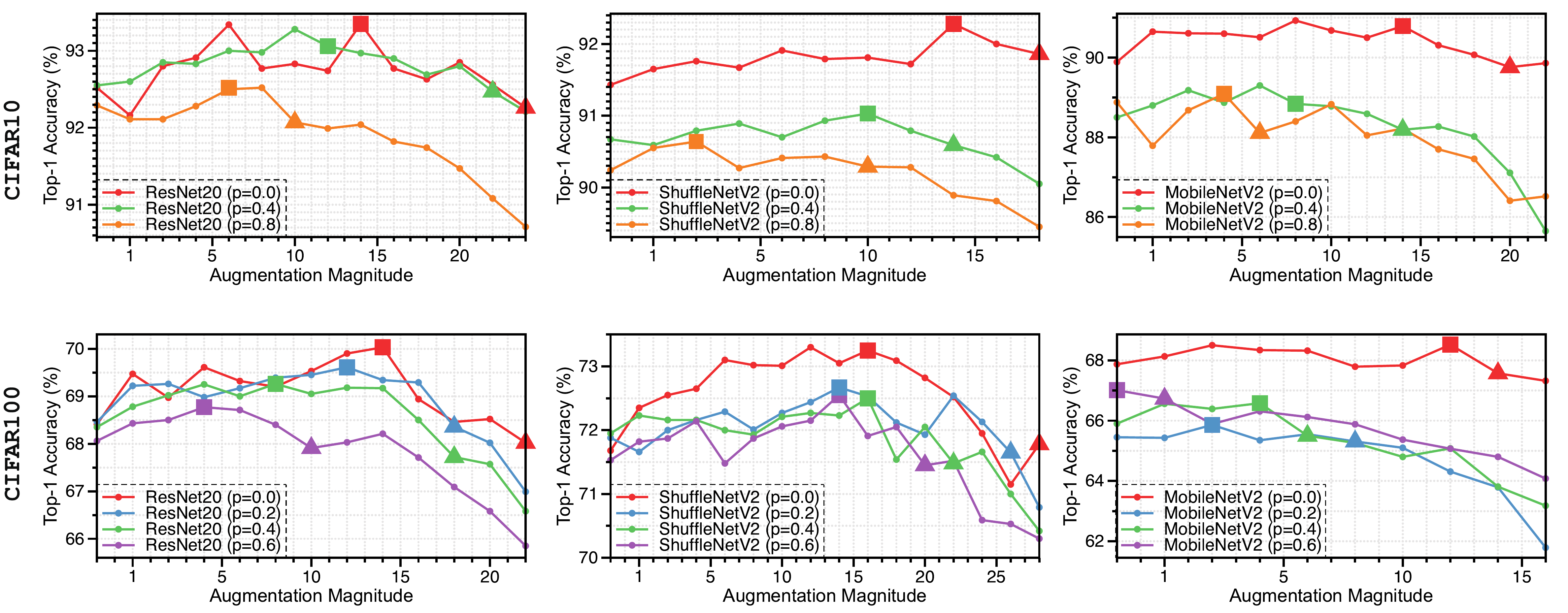}
    \caption{Experimental results of the pruned ResNet20, MobileNetV2, ShuffleNetV2 trained with RandAugment of different magnitudes on CIFAR10 and CIFAR100. The squares and triangles indicate the optimal magnitude and the maximal magnitude, respectively. $p$ indicates the pruning ratio. For instance, $p$=0.2 indicates 20\% parameters have been pruned to zero.}
    \label{fig:cifar10}
\end{center}
\end{figure*}

\section{Experiment}

\subsection{Models in Different Sizes Prefer Data Augmentation with Different Magnitudes\label{model_in_diff}}

\begin{figure}
\begin{center}
\vspace{-0.6cm}
\includegraphics[width=0.32\textwidth]{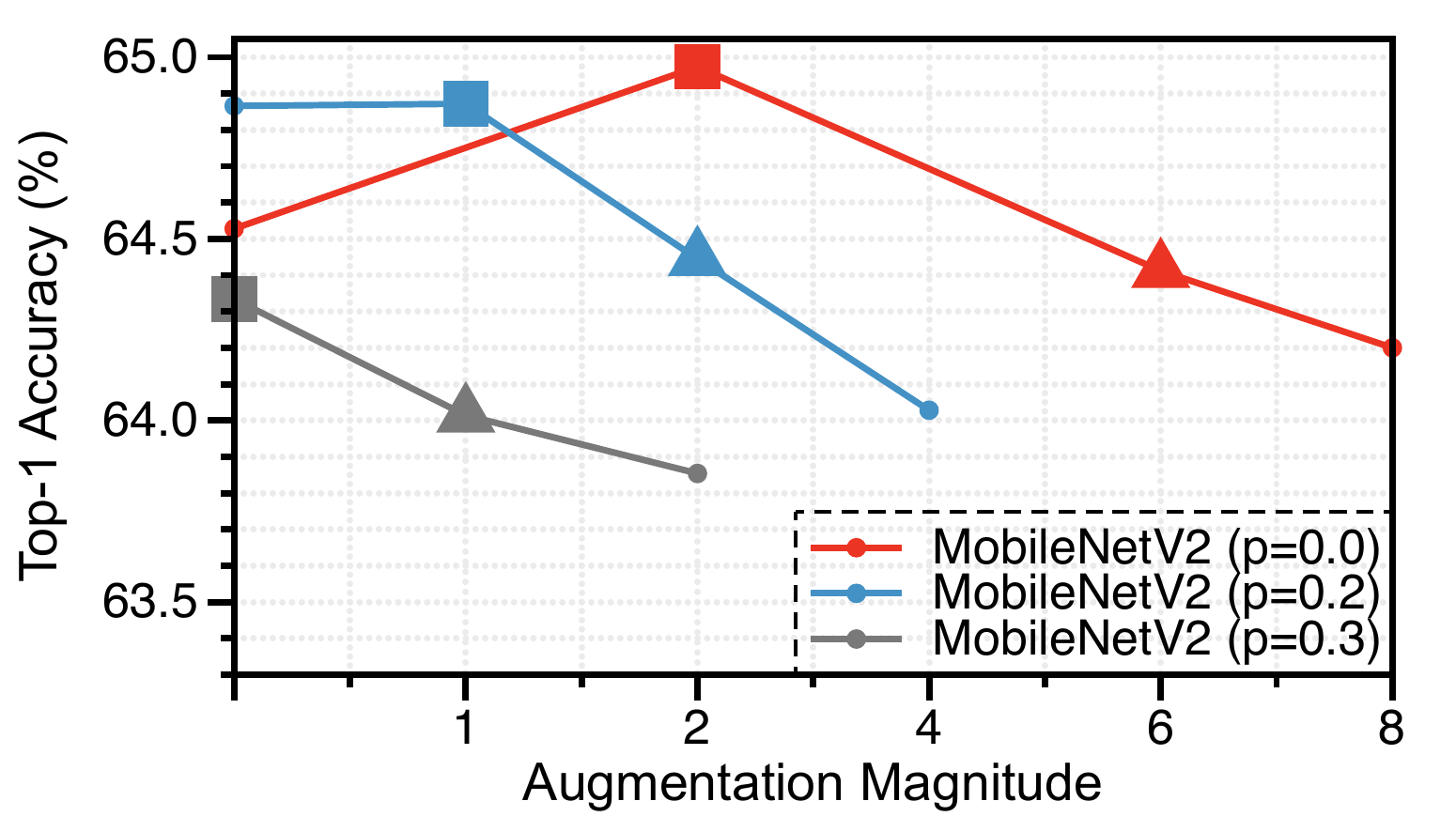}
\end{center}
\vspace{-0.2cm}
\caption{\label{fig:imagenet} Experiments of MobileNetV2 on ImageNet trained with data augmentation of different magnitudes.}
\vspace{-0.4cm}
\end{figure}
In this subsection, we show that there is a strong correlation between the sizes of models and their corresponding optimal magnitudes of data augmentation. Fig.~\ref{fig:cifar10} and Fig.~\ref{fig:imagenet} show the experimental results of neural networks of different pruning ratio trained with data augmentation of different magnitudes on CIFAR and ImageNet, respectively. Given a model with a specific pruning ratio, we denote its corresponding ``optimal magnitude'' as the magnitude which leads to the best performance and the ``maximal magnitude'' as the maximal magnitude of data augmentation which does not harm model performance compared with not using any data augmentation. Our experimental results demonstrate that:

    \textbf{(A)} A data augmentation with a very low magnitude tends to lead to limited performance improvements. In contrast, an over-large magnitude can harm model performance by a large margin. The optimal magnitude is usually a moderate value which introduces enough but not too much learning difficulty. For instance, on CIFAR100 with the unpruned ResNet20, the optimal data augmentation (M=14) leads to around 1.5\% accuracy improvements, while the lowest and the highest magnitudes lead to around 1.0\% and -0.5\% accuracy changes, respectively. 
    
    \textbf{(B)} A model in a larger size prefers to data augmentation with a higher magnitude. There is a strong correlation between the pruning ratio and the corresponding optimal and maximal magnitudes. For instance, on CIFAR100 with ResNet20, the optimal magnitude for the 60\% pruning, 40\% pruning, 20\% pruning, and unpruned models are 4, 8, 12, and 14, respectively.
    We suggest that this is because a model with more parameters has more representation ability to learn the strong regularization effect from  data augmentations with high magnitudes. In other words, the regularization effect from the strong data augmentation may be too challenging to be learned by a small model and thus harms its performance.
    For instance, On CIFAR100 with ResNet20, around 2\% accuracy drop can be observed when the data augmentation magnitude is 22.
    
    \textbf{(C)} When the neural networks have enough parameters, a data augmentation with a higher magnitude tends to lead to more significant accuracy boosts. For instance, on CIFAR100, the data augmentation with a higher magnitude (M=14) leads to around 1.5 accuracy boosts, while the data augmentation with a lower magnitude (M=1) leads to only around 0.6 accuracy boosts.
    This observation indicates that compared with a weak data augmentation, there is more potential knowledge in strong data augmentations, which can only be learned by models with enough representation ability.

 
\subsection{Pruning with Data Augmentation Magnitude Decay}
\begin{figure}
\begin{center}
\vspace{-0.6cm}
\includegraphics[width=0.35\textwidth]{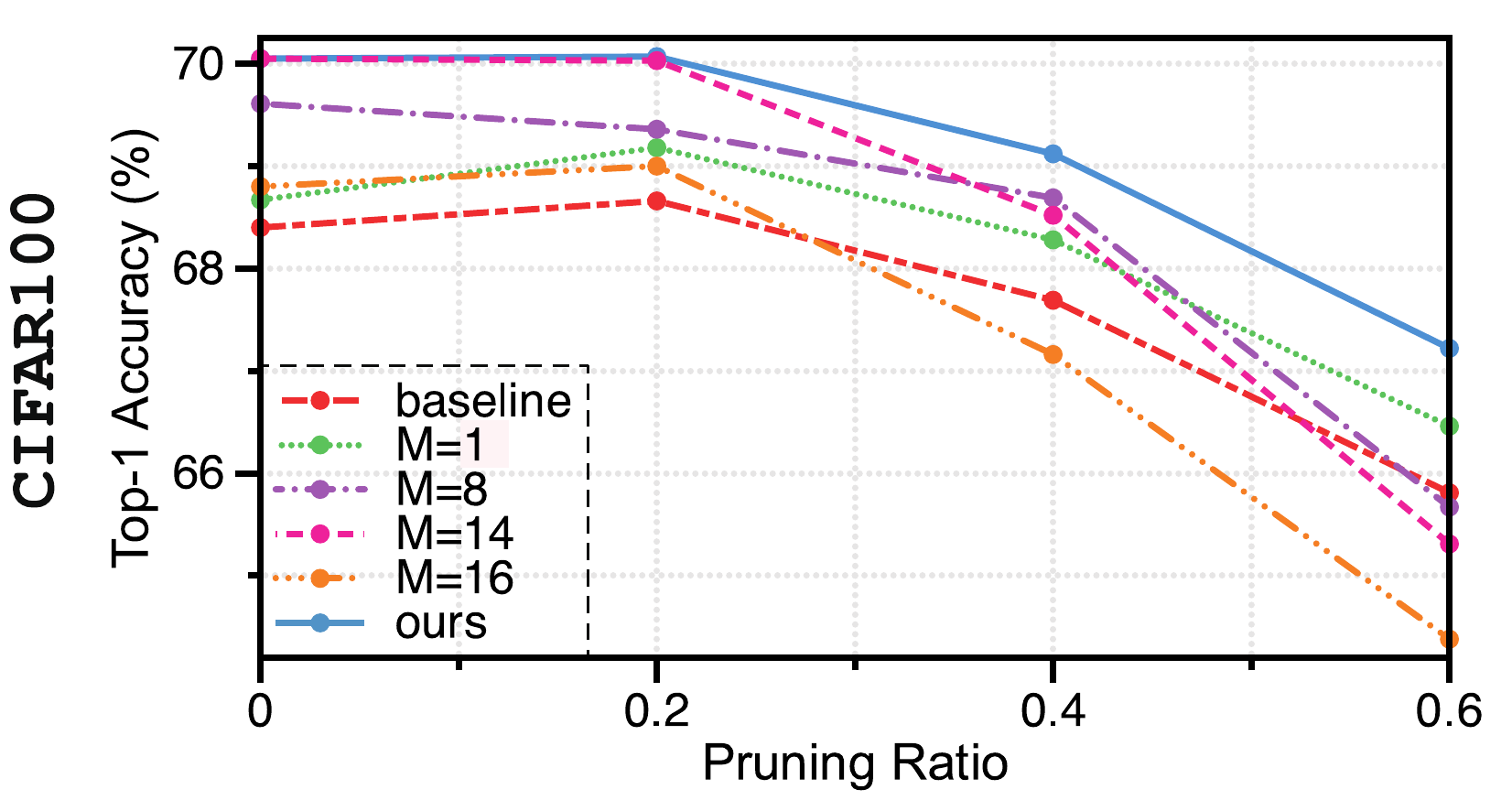}
\end{center}
\vspace{-0.2cm}
\caption{\label{fig:compare} Comparison between using the consistent data augmentation magnitude and decayed data augmentation magnitudes during pruning on CIFAR100.}
\vspace{-0.4cm}
\end{figure}
In most previous pruning research, the same data augmentation magnitude is utilized for neural networks of different pruning ratios. However, our previous observations demonstrate that models with different pruning ratios prefer data augmentation with different magnitudes. This observation motivates us to gradually reduce the data augmentation magnitude during pruning. Fig.~\ref{fig:compare} shows the experimental results of using the consistent or decayed magnitudes on CIFAR100 with ResNet20. \emph{baseline} indicates no data augmentation is applied. \emph{M} indicates the magnitude for data augmentation. 
It is observed that: \textbf{(A)} Compared with consistently using the four different magnitudes during the whole pruning period, gradually reducing the data augmentation magnitudes leads to significantly higher accuracy for models with different pruning ratios. \textbf{(B)} When data augmentation with a consistent magnitude is utilized for the whole pruning period, a higher magnitude tends to improve model performance in a low pruning ratio but reduces model performance in a high pruning ratio, and vice versa. We argue that this is because a higher magnitude leads to more benefits when the model has enough parameters but harms model performance when most of the parameters have been pruned.

\begin{figure*}[t]
    \centering
    \includegraphics[width=\linewidth]{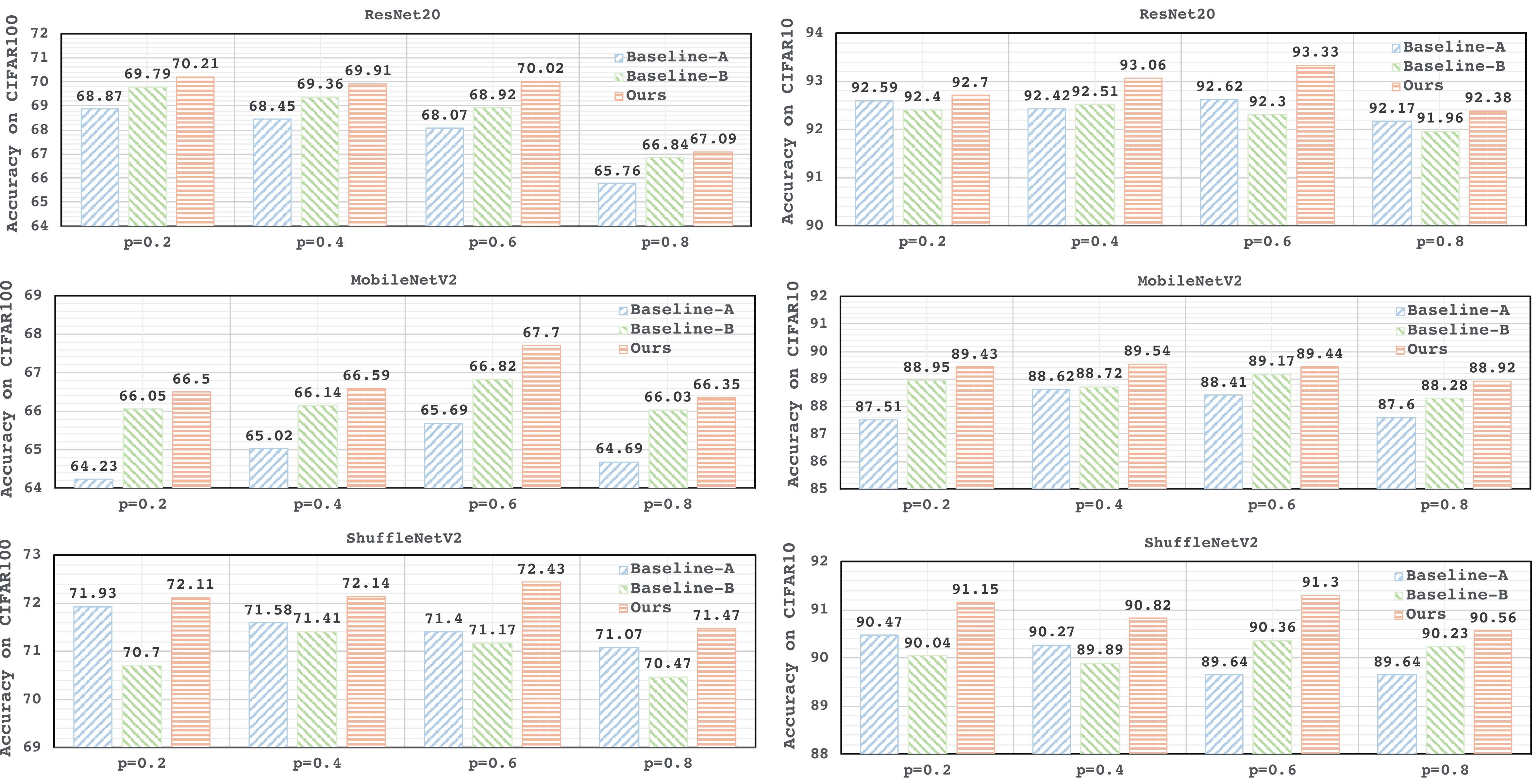}
    \caption{Experimental results of three augmentation schemes in pruning settings on CIFAR10 and CIFAR100 with ResNet20, MobileNetV2 and ShuffleNetV2. \emph{Baseline-A}: A pruned model is firstly trained with a strong data augmentation and then trained with a weak data augmentation (its corresponding optimal data augmentation). \emph{Baseline-B}: A pre-pruning model is firstly trained with a weak data augmentation, and then pruned and retrained with the weak data augmentation. \emph{Inhering Scheme (Ours)}: A pre-pruning model is firstly trained with a strong data augmentation, and then pruned and retrained with weak data augmentation (Fig.~\ref{fig:addition}-a). 
    }
    \label{fig:resnet20}
\end{figure*}
\subsection{Small Models Can Benefit from Strong Data Augmentations Indirectly}
\subsubsection{Inheriting the Knowledge of Strong Data Augmentations Learned Before Pruning\label{inherit_prune}}
\begin{figure*}
    \centering
    \includegraphics[width=\linewidth]{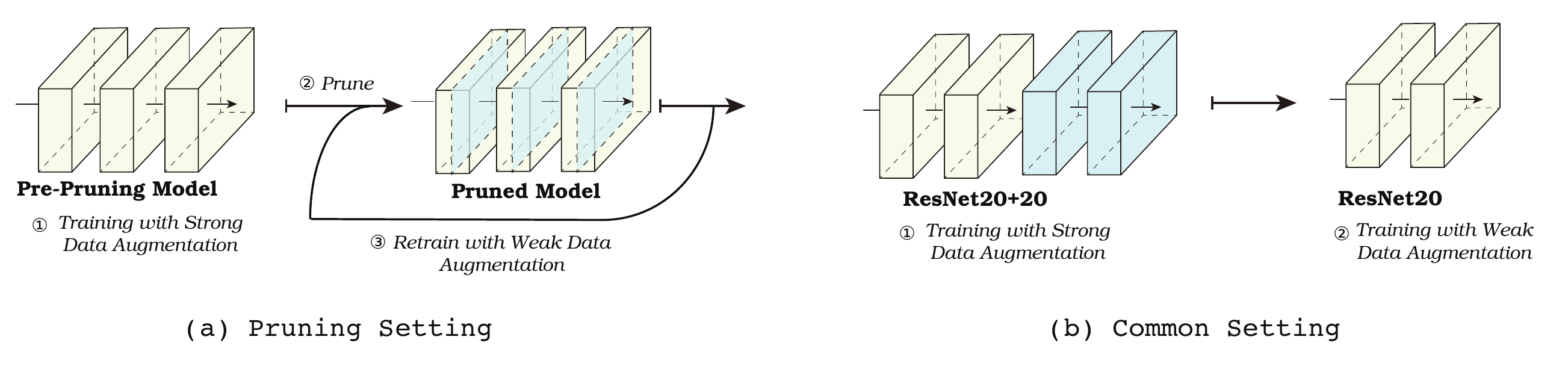}
    \caption{The overview of a small model indirectly learn knowledge of strong data augmentations in the pruning setting and the common setting. In the pruning setting, the pruned model can inherit the knowledge of strong data augmentations learned before pruning. In the common setting, the pruned model can inherit the knowledge of strong data augmentations learned with the additional layers.}
    \label{fig:addition}
\end{figure*}

Our previous experimental results show that it is challenging for a small model to directly learn with a strong data augmentation. In this subsection, we study whether a small model can benefit from strong data augmentations by inheriting the knowledge from the weights of a big model. Concretely, we have conducted the following three experiments: \emph{Baseline-A}: A pruned model is firstly trained with a strong data augmentation, and then trained with a weak data augmentation (its corresponding optimal data augmentation). \emph{Baseline-B}: A pre-pruning model is firstly trained with a weak data augmentation, and then pruned and retrained with the weak data augmentation. \emph{Our Scheme}: As shown in Fig.~\ref{fig:addition}(a), a pre-pruning model is firstly trained with a strong data augmentation, and then pruned and retrained with a weak data augmentation. Note that 
our scheme and the two baseline schemes finally obtain pruned network retrained with the same weak data augmentation in the same pruning ratios. Their main difference is that in their first training period: (a) Our method employs \emph{large} model (pre-pruning model) to learn the \emph{hard} data augmentation. (b) Baseline-A employs a \emph{small} model (pruned model) to learn the \emph{hard} data augmentation. (c) Baseline-B employs a \emph{small} model (pruned model) to learn the \emph{weak} data augmentation. 

Experimental results of the three schemes are shown in Fig.~\ref{fig:resnet20}. It is observed that our scheme achieves the highest accuracy in all the datasets, neural networks and pruning ratios. The superiority of our scheme over Baseline-A indicates that the knowledge from hard data augmentation can be learned from a large model (pre-pruning model) and then inherited by the small model (pruned model) during pruning. Besides, its superiority over Baseline-B confirms that the performance improvements in our scheme come from the usage of hard data augmentation instead of the two-stage training in pruning, and a small (pruned) model can not directly learn from a strong data augmentation.
These observations indicate that although directly applying a strong data augmentation to a small (pruned ) model harms its performance, the small (pruned) model can still benefit from strong data augmentations by firstly learning strong data augmentations with additional parameters and then inheriting them during neural network pruning.

\begin{figure*}
    \centering
    \includegraphics[width=\linewidth]{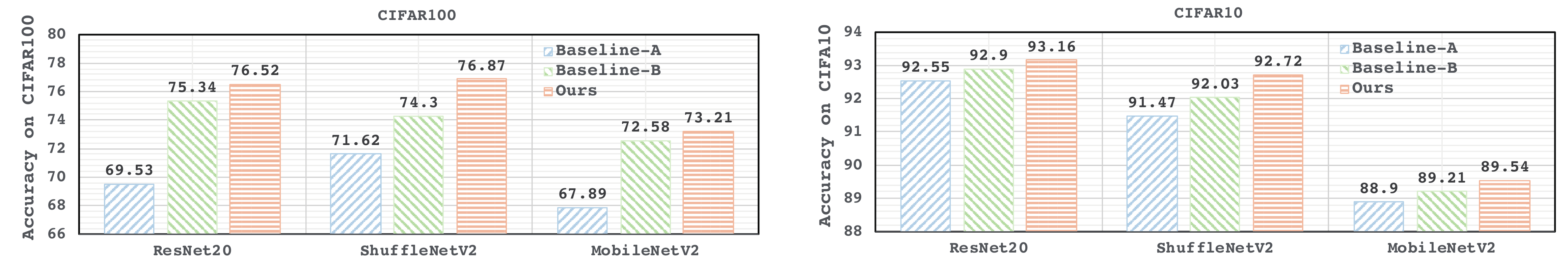}
    \caption{Experimental results of training small models to learn strong data augmentations with additional layers on CIFAR100 and CIFAR10 with ResNet20, MobileNetV2 and ShuffleNetV2. \emph{Baseline-A}: The small model is firstly trained with a strong data augmentation, and then trained with a weak data augmentation. \emph{Baseline-B}: The model is firstly trained with a weak data augmentation with additional layers and then discards the additional layers and retrained with the weak data augmentation. \emph{Inheriting Scheme (Ours)}: The model is firstly trained with a strong data augmentation with additional layers, and then discards additional layers and retrained with a weak data augmentation. }
    \label{fig:addition_exp}
\end{figure*}

\subsubsection{Inheriting Knowledge of Strong Data Augmentations Learned with Additional Layers}
In the last section, we show that a pruned model can inherit the knowledge of strong data augmentations learned by the pre-pruning model.
Motivated by its effectiveness, we further propose to extend this idea to the common neural network training settings beyond pruning. As shown in Fig.~\ref{fig:addition}(b), it consists of a two-stage training pipeline. 
Taking ResNet20 as an example, during the first training stage, additional several (\emph{e.g. 20}) layers are attached after the origin ResNet20 to improve its representation ability, and the obtained ResNet20+20 is trained with a strong data augmentation. In the second training stage, the additional 20 layers are discarded from the ResNet20+20, and the obtained ResNet20 is further trained with a weak data augmentation. Hence, the knowledge from a strong data augmentation can be firstly learned by the ResNet20+20 and then inherited by the ResNet20. 

Similar to the last section, we compare our scheme with the following two baselines. \emph{Baseline-A}: The small model is firstly trained with a strong data augmentation, and then trained with a weak data augmentation. \emph{Baseline-B}: The model is firstly trained with a weak data augmentation with additional layers, and then discards the additional layers and retrained with the weak data augmentation. Experimental results of ResNet20, MobileNetV2 and ShuffleNetV2 on CIFAR10 and CIFAR100 are shown in Fig.~\ref{fig:addition_exp}. It is observed that our scheme leads to significant accuracy improvements over the other two baselines, indicating the knowledge from a strong data augmentation learned by the additional layers can be inherited. Besides, Baseline-B consistently outperforms Baseline-A in all datasets and neural networks, indicating even without  strong data augmentations, learning with additional layers can also boost the performance of neural networks.

\begin{figure*}
    \centering
    \includegraphics[width=\linewidth]{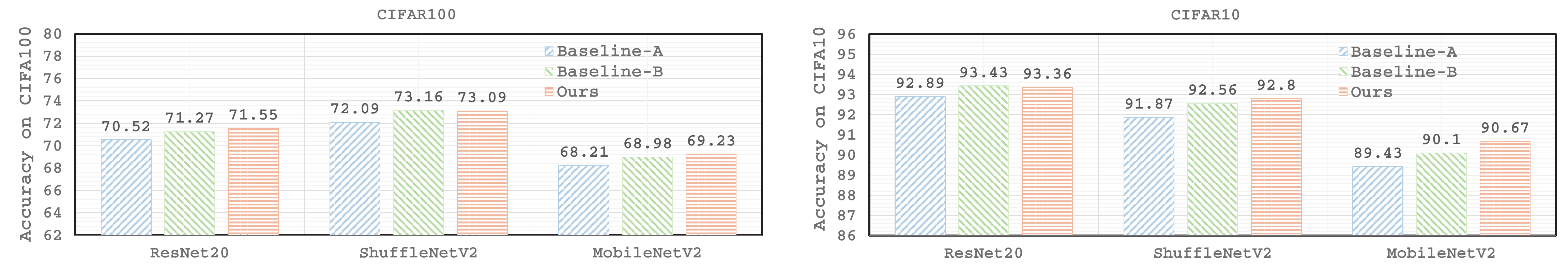}
    \caption{Experimental results of training small models (students) by employing pre-trained large models (teachers) to filter the hard data augmentation. Experiments are conducted with Hinton knowledge distillation~\cite{distill_hinton} with a ResNet110 teacher. \emph{Baseline-A}: Students are trained with data augmentation of random magnitudes. \emph{Baseline-B}: Students are trained with data augmentation of their optimal magnitude. \emph{Our scheme}: Students are trained with the data augmentation which can minimize Equation \ref{target}. Note that the optimal magnitudes utilized in Baseline-B are selected based on 20 experiments. Baseline-A and our scheme do not require experiments to select optimal magnitudes.} 
    \label{fig:filter}
\end{figure*}

\subsection{Employing Pre-trained Large Models to Filter Hard Augmentation for Small Models}
Our previous experiments show that the optimal data augmentation for a large model usually tends to be too strong for a small model. This observation intuitively suggests that given an augmented training sample, if a pre-trained large model can not predict it correctly, it has a high possibility to be a hard augmentation\footnote{Here we denote ``hard augmentation'' as the augmentation which is too hard to be learned by neural networks.} for a small model. This observation indicates that in knowledge distillation, the prediction of the pre-trained large teacher model can be utilized to filter the training samples with hard data augmentation to facilitate the training of the student model. Concretely, for each training sample $x$, we generate its augmentation by applying RandAugment with random magnitudes by $n$ times. Denoting its augmentation as $\{x_1, x_2, ..., x_n\}$, then, instead of distilling knowledge on all of them, we select one of them $x^*$ for student training, which can be formulated as 
\begin{equation}\label{target}
    x^* = \mathop{\arg\min}_{x_i} \alpha \cdot \mathcal{L}_{\text{CE}}(f_{\mathcal{T}}(x_i), y_i) - \beta \cdot \mathcal{L}_{\text{KL}}(f_{\mathcal{T}}(x_i), f_{\mathcal{S}}(x_i)),
\end{equation}
where the first item indicates the cross-entropy loss between teacher prediction and the label, and the second item indicates the KL divergence (\emph{i.e.,} knowledge distillation loss) between teacher prediction and student prediction.  As shown in Equation \ref{target}, the selected data augmentation $x^*$ has the following two properties. 
Firstly, it maximizes the difference between the prediction of the student and the teacher. In knowledge distillation, the student model is trained to minimize this distance. Thus, the data augmentation which can maximize this distance usually tends to have more value to be learned in knowledge distillation. Secondly, it minimizes the cross-entropy loss between the teacher prediction and the label, which limits that $x^*$ can be correctly predicted by the teacher, indicating $x^*$ should not be too hard to be learned by the student. With the two targets, our method can select the most valuable data augmentation from the randomly generated $n$ data augmentations and thus improve the efficiency of knowledge distillation.

Fig.~\ref{fig:filter} shows the experimental results of our scheme and two baselines. \emph{Baseline-A}: Students are trained with data augmentation of random magnitudes. \emph{Baseline-B}: Students are trained with data augmentation of their optimal magnitude. It is observed that our method outperforms the Baseline-A by a large margin, indicating that it can successfully select the most valuable data augmentation for knowledge distillation. Besides, in 4 of the 6 experiments, our scheme outperforms Baseline-B which uses data augmentation with optimal magnitude. We argue that this is because the used augmentation in our scheme is selected from multiple data augmentation with random magnitudes, and thus it leads to a larger data diversity than using the consistent data augmentation magnitude.  

\textbf{Ablation Study} The criterion of selecting data augmentation in Equation \ref{target} includes two targets: (a) minimizing the loss between teacher prediction to filter the hard data augmentation, and (b) maximizing the knowledge distillation loss to select the data augmentation which is more valuable for knowledge distillation. This paragraph gives the ablation study of the two targets. On CIFAR100 with ResNet20, our experimental results show that compared with Baseline-A (70.52\%), 0.59\%  and -1.24\% accuracy changes can be observed by only using the  (a) and (b) for augmentation selection, respectively. This observation indicates that by only using target (b) (\emph{i.e., } $\alpha=0$), the hard data augmentation may be selected for training and thus the performance of the student is harmed. By combining  (b) with (a), the hard data augmentation which can not be corrected predicted by the teacher model will not be selected and utilized in student training, and thus it achieves higher accuracy.

\vspace{-0.15cm}
\section{Discussion}

\vspace{-0.15cm}

\textbf{Rethinking the Value of Pruning} In a typical neural network pruning algorithm, the pruned neural network is usually finetuned (\emph{a.k.a.} retrained) based on their weights before training. Surprisingly, Liu~\emph{et al.} show that fine-tuning a pruned model only gives comparable or even worse performance than directly training the pruned model with randomly initialized weights~\cite{rethink}. However, all of the previous studies ignore the usage of data augmentation. As shown in our experiments in Section~\ref{inherit_prune}, the knowledge of the strong data augmentation learned by the pre-pruned models can be preserved to the pruned models by inheriting their weights. In contrast, directly training pruned models with randomly initialized weights with strong data augmentation does not improve but harm model performance. These observations indicate that the problem of whether finetuning a pruned model or training the pruned model with randomly initialized weights is better may have different answers when data augmentation of different magnitudes is utilized.
We hope this observation may promote research on rethinking the value of pruning in a more complex setting.

\textbf{Adaptive Data Augmentation or Consistent Data Augmentation} Previous data augmentation methods usually give a consist augmentation policy in different tasks. Recently, automatic data augmentation methods have been proposed to search the optimal data augmentation policy for the given dataset. However, as shown in our experiments of Section~\ref{model_in_diff}, different neural networks and even the same neural network in different pruning ratios prefer data augmentation with different magnitudes. These observations indicate that the adaptive data augmentation methods which can apply different augmentation policies to different datasets and models should be studied.

\vspace{-0.1cm}
\section{Conclusion}
\vspace{-0.1cm}
Instead of proposing new model compression or data augmentation methods, this paper gives an empirical and comprehensive study of data augmentation for model compression methods. In summary, we mainly have the following conclusions. \textbf{(A)} Models in different sizes prefer data augmentation with different magnitudes. Usually, a higher magnitude data augmentation significantly improves the performance of a large (pre-pruning) model, but harms the performance of a small (post-pruning) model. \textbf{(B)} A small (pruned) model still can benefit from hard data augmentation by firstly learning the hard data augmentation with additional parameters and then discarding them. The knowledge learned from hard data augmentation can still be inherited by the small (pruned) model even if the additional parameters have been discarded. \textbf{(C)}  The large model can be utilized to filter the hard data augmentation samples for the small model, which can efficiently find the optimal data augmentation policy for the small model. This observation may provide insightful guidance in designing efficient and effective data augmentation methods.

{\small
\bibliographystyle{plain}
\bibliography{egbib}

\begin{thebibliography}{100}

\bibitem{kd_detection4}
Mohammad~Farhadi Bajestani and Yezhou Yang.
\newblock Tkd: Temporal knowledge distillation for active perception.
\newblock In {\em The IEEE Winter Conference on Applications of Computer
  Vision}, pages 953--962, 2020.

\bibitem{yolact}
Daniel Bolya, Chong Zhou, Fanyi Xiao, and Yong~Jae Lee.
\newblock Yolact: Real-time instance segmentation.
\newblock In {\em Proceedings of the IEEE international conference on computer
  vision}, pages 9157--9166, 2019.

\bibitem{detect_prune2}
Maxim Bonnaerens, Matthias Freiberger, and Joni Dambre.
\newblock Anchor pruning for object detection.
\newblock {\em arXiv preprint arXiv:2104.00432}, 2021.

\bibitem{model_compression}
Cristian Buciluǎ, Rich Caruana, and Alexandru Niculescu-Mizil.
\newblock Model compression.
\newblock In {\em Proceedings of the 12th ACM SIGKDD international conference
  on Knowledge discovery and data mining}, pages 535--541. ACM, 2006.

\bibitem{kd_detection1}
Guobin Chen, Wongun Choi, Xiang Yu, Tony Han, and Manmohan Chandraker.
\newblock Learning efficient object detection models with knowledge
  distillation.
\newblock In {\em Advances in Neural Information Processing Systems}, pages
  742--751, 2017.

\bibitem{addernet}
Hanting Chen, Yunhe Wang, Chunjing Xu, Boxin Shi, Chao Xu, Qi~Tian, and Chang
  Xu.
\newblock Addernet: Do we really need multiplications in deep learning?
\newblock In {\em Proceedings of the IEEE/CVF Conference on Computer Vision and
  Pattern Recognition}, pages 1468--1477, 2020.

\bibitem{vit_prune2}
Tianlong Chen, Yu~Cheng, Zhe Gan, Lu~Yuan, Lei Zhang, and Zhangyang Wang.
\newblock Chasing sparsity in vision transformers: An end-to-end exploration.
\newblock {\em Advances in Neural Information Processing Systems}, 34, 2021.

\bibitem{simclr}
Ting Chen, Simon Kornblith, Mohammad Norouzi, and Geoffrey Hinton.
\newblock A simple framework for contrastive learning of visual
  representations.
\newblock {\em arXiv preprint arXiv:2002.05709}, 2020.

\bibitem{segment_prune1}
Xinghao Chen, Yunhe Wang, Yiman Zhang, Peng Du, Chunjing Xu, and Chang Xu.
\newblock Multi-task pruning for semantic segmentation networks.
\newblock {\em arXiv preprint arXiv:2007.08386}, 2020.

\bibitem{actication_quantization}
Jungwook Choi, Zhuo Wang, Swagath Venkataramani, Pierce I-Jen Chuang,
  Vijayalakshmi Srinivasan, and Kailash Gopalakrishnan.
\newblock Pact: Parameterized clipping activation for quantized neural
  networks.
\newblock {\em arXiv preprint arXiv:1805.06085}, 2018.

\bibitem{rand}
Ekin~D Cubuk, Barret Zoph, Jonathon Shlens, and Quoc~V Le.
\newblock Randaugment: Practical automated data augmentation with a reduced
  search space.
\newblock In {\em Proceedings of the IEEE/CVF Conference on Computer Vision and
  Pattern Recognition Workshops}, pages 702--703, 2020.

\bibitem{AutoAugmentLA}
Ekin~Dogus Cubuk, Barret Zoph, Dandelion Man{\'e}, Vijay Vasudevan, and Quoc~V.
  Le.
\newblock Autoaugment: Learning augmentation policies from data.
\newblock {\em ArXiv}, abs/1805.09501, 2018.

\bibitem{das2020empirical}
Deepan Das, Haley Massa, Abhimanyu Kulkarni, and Theodoros Rekatsinas.
\newblock An empirical analysis of the impact of data augmentation on knowledge
  distillation.
\newblock {\em arXiv preprint arXiv:2006.03810}, 2020.

\bibitem{bert}
Jacob Devlin, Ming-Wei Chang, Kenton Lee, and Kristina Toutanova.
\newblock Bert: Pre-training of deep bidirectional transformers for language
  understanding.
\newblock In {\em NAACL}, 2018.

\bibitem{cutout}
Terrance DeVries and Graham~W Taylor.
\newblock Improved regularization of convolutional neural networks with cutout.
\newblock {\em arXiv preprint arXiv:1708.04552}, 2017.

\bibitem{vit}
Alexey Dosovitskiy, Lucas Beyer, Alexander Kolesnikov, Dirk Weissenborn,
  Xiaohua Zhai, Thomas Unterthiner, Mostafa Dehghani, Matthias Minderer, Georg
  Heigold, Sylvain Gelly, et~al.
\newblock An image is worth 16x16 words: Transformers for image recognition at
  scale.
\newblock {\em arXiv preprint arXiv:2010.11929}, 2020.

\bibitem{kd_aug}
Jie Fu, Xue Geng, Zhijian Duan, Bohan Zhuang, Xingdi Yuan, Adam Trischler, Jie
  Lin, Chris Pal, and Hao Dong.
\newblock Role-wise data augmentation for knowledge distillation.
\newblock {\em arXiv preprint arXiv:2004.08861}, 2020.

\bibitem{fu2020role}
Jie Fu, Xue Geng, Zhijian Duan, Bohan Zhuang, Xingdi Yuan, Adam Trischler, Jie
  Lin, Chris Pal, and Hao Dong.
\newblock Role-wise data augmentation for knowledge distillation.
\newblock {\em arXiv preprint arXiv:2004.08861}, 2020.

\bibitem{detect_prune3}
Sanjukta Ghosh, Shashi~KK Srinivasa, Peter Amon, Andreas Hutter, and Andr{\'e}
  Kaup.
\newblock Deep network pruning for object detection.
\newblock In {\em 2019 IEEE International Conference on Image Processing
  (ICIP)}, pages 3915--3919. IEEE, 2019.

\bibitem{deepcompression}
Song Han, Huizi Mao, and William~J Dally.
\newblock Deep compression: Compressing deep neural networks with pruning,
  trained quantization and huffman coding.
\newblock In {\em ICLR}, 2016.

\bibitem{hassibi1993second}
Babak Hassibi and David~G Stork.
\newblock Second order derivatives for network pruning: Optimal brain surgeon.
\newblock In {\em Advances in neural information processing systems}, pages
  164--171, 1993.

\bibitem{fasteraug}
Ryuichiro Hataya, Jan Zdenek, Kazuki Yoshizoe, and Hideki Nakayama.
\newblock Faster autoaugment: Learning augmentation strategies using
  backpropagation.
\newblock In {\em European Conference on Computer Vision}, pages 1--16.
  Springer, 2020.

\bibitem{moco}
Kaiming He, Haoqi Fan, Yuxin Wu, Saining Xie, and Ross Girshick.
\newblock Momentum contrast for unsupervised visual representation learning.
\newblock In {\em Proceedings of the IEEE/CVF Conference on Computer Vision and
  Pattern Recognition}, pages 9729--9738, 2020.

\bibitem{resnet}
Kaiming He, Xiangyu Zhang, Shaoqing Ren, and Jian Sun.
\newblock Deep residual learning for image recognition.
\newblock In {\em CVPR}, pages 770--778, 2016.

\bibitem{segment_prune2}
Wei He, Meiqing Wu, Mingfu Liang, and Siew-Kei Lam.
\newblock Cap: Context-aware pruning for semantic segmentation.
\newblock In {\em Proceedings of the IEEE/CVF Winter Conference on Applications
  of Computer Vision}, pages 960--969, 2021.

\bibitem{filter_prune}
Yang He, Guoliang Kang, Xuanyi Dong, Yanwei Fu, and Yi~Yang.
\newblock Soft filter pruning for accelerating deep convolutional neural
  networks.
\newblock {\em arXiv preprint arXiv:1808.06866}, 2018.

\bibitem{geometric_prune}
Yang He, Ping Liu, Ziwei Wang, Zhilan Hu, and Yi~Yang.
\newblock Filter pruning via geometric median for deep convolutional neural
  networks acceleration.
\newblock In {\em Proceedings of the IEEE Conference on Computer Vision and
  Pattern Recognition}, pages 4340--4349, 2019.

\bibitem{channel_prune}
Yihui He, Xiangyu Zhang, and Jian Sun.
\newblock Channel pruning for accelerating very deep neural networks.
\newblock In {\em Proceedings of the IEEE International Conference on Computer
  Vision}, pages 1389--1397, 2017.

\bibitem{BenchmarkingNN_classifiction}
Dan Hendrycks and Thomas~G. Dietterich.
\newblock Benchmarking neural network robustness to common corruptions and
  perturbations.
\newblock {\em ArXiv}, abs/1903.12261, 2019.

\bibitem{kd_comprehensive}
Byeongho Heo, Jeesoo Kim, Sangdoo Yun, Hyojin Park, Nojun Kwak, and Jin~Young
  Choi.
\newblock A comprehensive overhaul of feature distillation.
\newblock In {\em Proceedings of the IEEE International Conference on Computer
  Vision}, pages 1921--1930, 2019.

\bibitem{kd_ab}
Byeongho Heo, Minsik Lee, Sangdoo Yun, and Jin~Young Choi.
\newblock Knowledge transfer via distillation of activation boundaries formed
  by hidden neurons.
\newblock In {\em Proceedings of the AAAI Conference on Artificial
  Intelligence}, volume~33, pages 3779--3787, 2019.

\bibitem{distill_hinton}
Geoffrey Hinton, Oriol Vinyals, and Jeff Dean.
\newblock Distilling the knowledge in a neural network.
\newblock In {\em NeurIPS}, 2014.

\bibitem{ho2019population}
Daniel Ho, Eric Liang, Xi~Chen, Ion Stoica, and Pieter Abbeel.
\newblock Population based augmentation: Efficient learning of augmentation
  policy schedules.
\newblock In {\em International Conference on Machine Learning}, pages
  2731--2741. PMLR, 2019.

\bibitem{mobilenet3}
Andrew Howard, Mark Sandler, Grace Chu, Liang-Chieh Chen, Bo~Chen, Mingxing
  Tan, Weijun Wang, Yukun Zhu, Ruoming Pang, Vijay Vasudevan, et~al.
\newblock Searching for mobilenetv3.
\newblock {\em arXiv preprint arXiv:1905.02244}, 2019.

\bibitem{mobilenets}
Andrew~G Howard, Menglong Zhu, Bo~Chen, Dmitry Kalenichenko, Weijun Wang,
  Tobias Weyand, Marco Andreetto, and Hartwig Adam.
\newblock Mobilenets: Efficient convolutional neural networks for mobile vision
  applications.
\newblock In {\em CVPR}, 2017.

\bibitem{3d_segment}
Qingyong Hu, Bo~Yang, Linhai Xie, Stefano Rosa, Yulan Guo, Zhihua Wang, Niki
  Trigoni, and Andrew Markham.
\newblock Randla-net: Efficient semantic segmentation of large-scale point
  clouds.
\newblock In {\em Proceedings of the IEEE/CVF Conference on Computer Vision and
  Pattern Recognition}, pages 11108--11117, 2020.

\bibitem{squeezenet}
Forrest~N Iandola, Song Han, Matthew~W Moskewicz, Khalid Ashraf, William~J
  Dally, and Kurt Keutzer.
\newblock Squeezenet: Alexnet-level accuracy with 50x fewer parameters and< 0.5
  mb model size.
\newblock In {\em ICLR}, 2016.

\bibitem{dataaug_robust}
Philip~TG Jackson, Amir~Atapour Abarghouei, Stephen Bonner, Toby~P Breckon, and
  Boguslaw Obara.
\newblock Style augmentation: data augmentation via style randomization.
\newblock In {\em CVPR Workshops}, volume~6, pages 10--11, 2019.

\bibitem{dataaug_fair}
Nikita Jaipuria, Xianling Zhang, Rohan Bhasin, Mayar Arafa, Punarjay
  Chakravarty, Shubham Shrivastava, Sagar Manglani, and Vidya~N Murali.
\newblock Deflating dataset bias using synthetic data augmentation.
\newblock In {\em Proceedings of the IEEE/CVF Conference on Computer Vision and
  Pattern Recognition Workshops}, pages 772--773, 2020.

\bibitem{teacher_do_more_gankd}
Qing Jin, Jian Ren, Oliver~J Woodford, Jiazhuo Wang, Geng Yuan, Yanzhi Wang,
  and Sergey Tulyakov.
\newblock Teachers do more than teach: Compressing image-to-image models.
\newblock In {\em Proceedings of the IEEE/CVF Conference on Computer Vision and
  Pattern Recognition}, pages 13600--13611, 2021.

\bibitem{Benchmarking_segmentation}
Christoph Kamann and Carsten Rother.
\newblock Benchmarking the robustness of semantic segmentation models.
\newblock {\em ArXiv}, abs/1908.05005, 2019.

\bibitem{NAS_kd}
Minsoo Kang, Jonghwan Mun, and Bohyung Han.
\newblock Towards oracle knowledge distillation with neural architecture
  search.
\newblock In {\em AAAI}, 2020.

\bibitem{dataaug_text2}
Sosuke Kobayashi.
\newblock Contextual augmentation: Data augmentation by words with paradigmatic
  relations.
\newblock {\em arXiv preprint arXiv:1805.06201}, 2018.

\bibitem{Alexnet}
Alex Krizhevsky, Ilya Sutskever, and Geoffrey~E Hinton.
\newblock Imagenet classification with deep convolutional neural networks.
\newblock In {\em NeurIPS}, pages 1097--1105, 2012.

\bibitem{optimal_brain_damage}
Yann LeCun, John~S Denker, and Sara~A Solla.
\newblock Optimal brain damage.
\newblock In {\em Advances in neural information processing systems}, pages
  598--605, 1990.

\bibitem{quantization_2}
Kibok Lee, Kimin Lee, Jinwoo Shin, and Honglak Lee.
\newblock Overcoming catastrophic forgetting with unlabeled data in the wild.
\newblock In {\em The IEEE International Conference on Computer Vision (ICCV)},
  October 2019.

\bibitem{gan_compress}
Muyang Li, Ji~Lin, Yaoyao Ding, Zhijian Liu, Jun-Yan Zhu, and Song Han.
\newblock Gan compression: Efficient architectures for interactive conditional
  gans.
\newblock In {\em Proceedings of the IEEE/CVF Conference on Computer Vision and
  Pattern Recognition}, pages 5284--5294, 2020.

\bibitem{kd_detection2}
Quanquan Li, Shengying Jin, and Junjie Yan.
\newblock Mimicking very efficient network for object detection.
\newblock In {\em Proceedings of the ieee conference on computer vision and
  pattern recognition}, pages 6356--6364, 2017.

\bibitem{revisit_discriminator}
Shaojie Li, Jie Wu, Xuefeng Xiao, Fei Chao, Xudong Mao, and Rongrong Ji.
\newblock Revisiting discriminator in {GAN} compression: {A}
  generator-discriminator cooperative compression scheme.
\newblock {\em CoRR}, abs/2110.14439, 2021.

\bibitem{local_kd}
Xiaojie Li, Jianlong Wu, Hongyu Fang, Yue Liao, Fei Wang, and Chen Qian.
\newblock Local correlation consistency for knowledge distillation.
\newblock In {\em European Conference on Computer Vision}, pages 18--33.
  Springer, 2020.

\bibitem{dada}
Yonggang Li, Guosheng Hu, Yongtao Wang, Timothy Hospedales, Neil~M Robertson,
  and Yongxin Yang.
\newblock Dada: differentiable automatic data augmentation.
\newblock {\em arXiv preprint arXiv:2003.03780}, 2020.

\bibitem{spkd_gan}
Zeqi Li, Ruowei Jiang, and Parham Aarabi.
\newblock Semantic relation preserving knowledge distillation for
  image-to-image translation.
\newblock In {\em European Conference on Computer Vision}, pages 648--663.
  Springer, 2020.

\bibitem{ohl}
Chen Lin, Minghao Guo, Chuming Li, Xin Yuan, Wei Wu, Junjie Yan, Dahua Lin, and
  Wanli Ouyang.
\newblock Online hyper-parameter learning for auto-augmentation strategy.
\newblock In {\em Proceedings of the IEEE/CVF International Conference on
  Computer Vision}, pages 6579--6588, 2019.

\bibitem{lin2020weight}
Ye~Lin, Yanyang Li, Ziyang Wang, Bei Li, Quan Du, Tong Xiao, and Jingbo Zhu.
\newblock Weight distillation: Transferring the knowledge in neural network
  parameters.
\newblock {\em arXiv preprint arXiv:2009.09152}, 2020.

\bibitem{ddas}
Aoming Liu, Zehao Huang, Zhiwu Huang, and Naiyan Wang.
\newblock Direct differentiable augmentation search.
\newblock In {\em Proceedings of the IEEE/CVF International Conference on
  Computer Vision}, pages 12219--12228, 2021.

\bibitem{structured_kd}
Yifan Liu, Ke~Chen, Chris Liu, Zengchang Qin, Zhenbo Luo, and Jingdong Wang.
\newblock Structured knowledge distillation for semantic segmentation.
\newblock In {\em Proceedings of the IEEE Conference on Computer Vision and
  Pattern Recognition}, pages 2604--2613, 2019.

\bibitem{metapruing}
Zechun Liu, Haoyuan Mu, Xiangyu Zhang, Zichao Guo, Xin Yang, Kwang-Ting Cheng,
  and Jian Sun.
\newblock Metapruning: Meta learning for automatic neural network channel
  pruning.
\newblock In {\em The IEEE International Conference on Computer Vision (ICCV)},
  October 2019.

\bibitem{rethink}
Zhuang Liu, Mingjie Sun, Tinghui Zhou, Gao Huang, and Trevor Darrell.
\newblock Rethinking the value of network pruning.
\newblock {\em arXiv preprint arXiv:1810.05270}, 2018.

\bibitem{pruing_l0}
Christos Louizos, Max Welling, and Diederik~P Kingma.
\newblock Learning sparse neural networks through $ l\_0 $ regularization.
\newblock {\em arXiv preprint arXiv:1712.01312}, 2017.

\bibitem{luo2017thinet}
Jian-Hao Luo, Jianxin Wu, and Weiyao Lin.
\newblock Thinet: A filter level pruning method for deep neural network
  compression.
\newblock In {\em Proceedings of the IEEE international conference on computer
  vision}, pages 5058--5066, 2017.

\bibitem{shufflenet}
Ningning Ma, Xiangyu Zhang, Hai-Tao Zheng, and Jian Sun.
\newblock Shufflenet v2: Practical guidelines for efficient cnn architecture
  design.
\newblock In {\em Proceedings of the European Conference on Computer Vision
  (ECCV)}, pages 116--131, 2018.

\bibitem{data_free_quantization}
Markus Nagel, Mart~van Baalen, Tijmen Blankevoort, and Max Welling.
\newblock Data-free quantization through weight equalization and bias
  correction.
\newblock In {\em The IEEE International Conference on Computer Vision (ICCV)},
  October 2019.

\bibitem{niu2020patdnn}
Wei Niu, Xiaolong Ma, Sheng Lin, Shihao Wang, Xuehai Qian, Xue Lin, Yanzhi
  Wang, and Bin Ren.
\newblock Patdnn: Achieving real-time dnn execution on mobile devices with
  pattern-based weight pruning.
\newblock In {\em Proceedings of the Twenty-Fifth International Conference on
  Architectural Support for Programming Languages and Operating Systems}, pages
  907--922, 2020.

\bibitem{kd_defense}
Nicolas Papernot, Patrick McDaniel, Xi~Wu, Somesh Jha, and Ananthram Swami.
\newblock Distillation as a defense to adversarial perturbations against deep
  neural networks.
\newblock In {\em 2016 IEEE Symposium on Security and Privacy (SP)}, pages
  582--597. IEEE, 2016.

\bibitem{relational_kd}
Wonpyo Park, Dongju Kim, Yan Lu, and Minsu Cho.
\newblock Relational knowledge distillation.
\newblock In {\em Proceedings of the IEEE Conference on Computer Vision and
  Pattern Recognition}, pages 3967--3976, 2019.

\bibitem{pytorch}
Adam Paszke, Sam Gross, Francisco Massa, Adam Lerer, James Bradbury, Gregory
  Chanan, Trevor Killeen, Zeming Lin, Natalia Gimelshein, Luca Antiga, Alban
  Desmaison, Andreas Kopf, Edward Yang, Zachary DeVito, Martin Raison, Alykhan
  Tejani, Sasank Chilamkurthy, Benoit Steiner, Lu~Fang, Junjie Bai, and Soumith
  Chintala.
\newblock Pytorch: An imperative style, high-performance deep learning library.
\newblock In {\em Advances in Neural Information Processing Systems 32}, pages
  8024--8035. Curran Associates, Inc., 2019.

\bibitem{thundernet}
Zheng Qin, Zeming Li, Zhaoning Zhang, Yiping Bao, Gang Yu, Yuxing Peng, and
  Jian Sun.
\newblock Thundernet: Towards real-time generic object detection on mobile
  devices.
\newblock In {\em The IEEE International Conference on Computer Vision (ICCV)},
  October 2019.

\bibitem{omgd}
Yuxi Ren, Jie Wu, Xuefeng Xiao, and Jianchao Yang.
\newblock Online multi-granularity distillation for gan compression.
\newblock {\em CoRR}, abs/2108.06908, 2021.

\bibitem{layer_prune}
Youngmin Ro and Jin~Young Choi.
\newblock Layer-wise pruning and auto-tuning of layer-wise learning rates in
  fine-tuning of deep networks.
\newblock {\em arXiv preprint arXiv:2002.06048}, 2020.

\bibitem{mobilenetv2}
Mark Sandler, Andrew Howard, Menglong Zhu, Andrey Zhmoginov, and Liang-Chieh
  Chen.
\newblock Mobilenetv2: Inverted residuals and linear bottlenecks.
\newblock In {\em Proceedings of the IEEE Conference on Computer Vision and
  Pattern Recognition}, pages 4510--4520, 2018.

\bibitem{kd_bert1}
Victor Sanh, Lysandre Debut, Julien Chaumond, and Thomas Wolf.
\newblock Distilbert, a distilled version of bert: smaller, faster, cheaper and
  lighter.
\newblock {\em arXiv preprint arXiv:1910.01108}, 2019.

\bibitem{dataaug_text1}
Connor Shorten, Taghi~M Khoshgoftaar, and Borko Furht.
\newblock Text data augmentation for deep learning.
\newblock {\em Journal of big Data}, 8(1):1--34, 2021.

\bibitem{song2020addersr}
Dehua Song, Yunhe Wang, Hanting Chen, Chang Xu, Chunjing Xu, and DaCheng Tao.
\newblock Addersr: Towards energy efficient image super-resolution.
\newblock {\em arXiv preprint arXiv:2009.08891}, 2020.

\bibitem{suzuki2022teachaugment}
Teppei Suzuki.
\newblock Teachaugment: Data augmentation optimization using teacher knowledge.
\newblock {\em arXiv preprint arXiv:2202.12513}, 2022.

\bibitem{efficientdet}
M~Tan, R~Pang, and QV~Le.
\newblock Efficientdet: Scalable and efficient object detection. arxiv 2019.
\newblock {\em arXiv preprint arXiv:1911.09070}, 2019.

\bibitem{kd_crd}
Yonglong Tian, Dilip Krishnan, and Phillip Isola.
\newblock Contrastive representation distillation.
\newblock {\em arXiv preprint arXiv:1910.10699}, 2019.

\bibitem{relational_kd2}
Frederick Tung and Greg Mori.
\newblock Similarity-preserving knowledge distillation.
\newblock In {\em Proceedings of the IEEE International Conference on Computer
  Vision}, pages 1365--1374, 2019.

\bibitem{wang2020gan}
Haotao Wang, Shupeng Gui, Haichuan Yang, Ji~Liu, and Zhangyang Wang.
\newblock Gan slimming: All-in-one gan compression by a unified optimization
  framework.
\newblock In {\em European Conference on Computer Vision}, pages 54--73.
  Springer, 2020.

\bibitem{wang2020knowledge}
Huan Wang, Suhas Lohit, Michael Jones, and Yun Fu.
\newblock Knowledge distillation thrives on data augmentation.
\newblock {\em arXiv preprint arXiv:2012.02909}, 2020.

\bibitem{lstm_prune1}
Shaorun Wang, Peng Lin, Ruihan Hu, Hao Wang, Jin He, Qijun Huang, and Sheng
  Chang.
\newblock Acceleration of lstm with structured pruning method on fpga.
\newblock {\em IEEE Access}, 7:62930--62937, 2019.

\bibitem{kd_detection3}
Tao Wang, Li~Yuan, Xiaopeng Zhang, and Jiashi Feng.
\newblock Distilling object detectors with fine-grained feature imitation.
\newblock In {\em Proceedings of the IEEE Conference on Computer Vision and
  Pattern Recognition}, pages 4933--4942, 2019.

\bibitem{dataaug_text3}
Jason Wei and Kai Zou.
\newblock Eda: Easy data augmentation techniques for boosting performance on
  text classification tasks.
\newblock {\em arXiv preprint arXiv:1901.11196}, 2019.

\bibitem{wei2020circumventing}
Longhui Wei, An~Xiao, Lingxi Xie, Xiaopeng Zhang, Xin Chen, and Qi~Tian.
\newblock Circumventing outliers of autoaugment with knowledge distillation.
\newblock In {\em European Conference on Computer Vision}, pages 608--625.
  Springer, 2020.

\bibitem{wu2020lite}
Zhanghao Wu, Zhijian Liu, Ji~Lin, Yujun Lin, and Song Han.
\newblock Lite transformer with long-short range attention.
\newblock {\em arXiv preprint arXiv:2004.11886}, 2020.

\bibitem{detect_prune1}
Zihao Xie, Li~Zhu, Lin Zhao, Bo~Tao, Liman Liu, and Wenbing Tao.
\newblock Localization-aware channel pruning for object detection.
\newblock {\em Neurocomputing}, 403:400--408, 2020.

\bibitem{kd_bert2}
Canwen Xu, Wangchunshu Zhou, Tao Ge, Furu Wei, and Ming Zhou.
\newblock Bert-of-theseus: Compressing bert by progressive module replacing.
\newblock {\em arXiv preprint arXiv:2002.02925}, 2020.

\bibitem{yang2019region}
Zhen Yang, Zhipeng Wang, Wenshan Xu, Xiuying He, Zhichao Wang, and Zhijian Yin.
\newblock Region-aware random erasing.
\newblock In {\em 2019 IEEE 19th International Conference on Communication
  Technology (ICCT)}, pages 1699--1703. IEEE, 2019.

\bibitem{adversarial_pruning}
Shaokai Ye, Kaidi Xu, Sijia Liu, Hao Cheng, Jan-Henrik Lambrechts, Huan Zhang,
  Aojun Zhou, Kaisheng Ma, Yanzhi Wang, and Xue Lin.
\newblock Adversarial robustness vs. model compression, or both?
\newblock In {\em The IEEE International Conference on Computer Vision (ICCV)},
  October 2019.

\bibitem{dataaug_superresolution}
Jaejun Yoo, Namhyuk Ahn, and Kyung-Ah Sohn.
\newblock Rethinking data augmentation for image super-resolution: A
  comprehensive analysis and a new strategy.
\newblock In {\em Proceedings of the IEEE/CVF Conference on Computer Vision and
  Pattern Recognition}, pages 8375--8384, 2020.

\bibitem{slimmablev2}
Jiahui Yu and Thomas~S. Huang.
\newblock Universally slimmable networks and improved training techniques.
\newblock In {\em The IEEE International Conference on Computer Vision (ICCV)},
  October 2019.

\bibitem{slimmable}
Jiahui Yu, Linjie Yang, Ning Xu, Jianchao Yang, and Thomas Huang.
\newblock Slimmable neural networks.
\newblock In {\em ICLR}, 2019.

\bibitem{attentiondistillation}
Sergey Zagoruyko and Nikos Komodakis.
\newblock Paying more attention to attention: Improving the performance of
  convolutional neural networks via attention transfer.
\newblock In {\em ICLR}, 2017.

\bibitem{zhang2017mixup}
Hongyi Zhang, Moustapha Cisse, Yann~N Dauphin, and David Lopez-Paz.
\newblock mixup: Beyond empirical risk minimization.
\newblock {\em arXiv preprint arXiv:1710.09412}, 2017.

\bibitem{wkd}
Linfeng Zhang, Xin Chen, Xiaobing Tu, Pengfei Wan, Ning Xu, and Kaisheng Ma.
\newblock Wavelet knowledge distillation: Towards efficient image-to-image
  translation.
\newblock In {\em CVPR}, 2022.

\bibitem{detectiondistillation}
Linfeng Zhang and Ma~Kaisheng.
\newblock Improve object detection with feature-based knowledge distillation:
  Towards accurate and efficient detectors.
\newblock In {\em ICLR}, 2021.

\bibitem{tofd}
Linfeng Zhang, Yukang Shi, Zuoqiang Shi, Kaisheng Ma, and Chenglong Bao.
\newblock Task-oriented feature distillation.
\newblock In {\em NeurIPS}, 2020.

\bibitem{selfdistillation}
Linfeng Zhang, Jiebo Song, Anni Gao, Jingwei Chen, Chenglong Bao, and Kaisheng
  Ma.
\newblock Be your own teacher: Improve the performance of convolutional neural
  networks via self distillation.
\newblock In {\em arXiv preprint:1905.08094}, 2019.

\bibitem{Zhang2019SCANAS}
Linfeng Zhang, Zhanhong Tan, Jiebo Song, Jingwei Chen, Chenglong Bao, and
  Kaisheng Ma.
\newblock Scan: A scalable neural networks framework towards compact and
  efficient models.
\newblock {\em ArXiv}, abs/1906.03951, 2019.

\bibitem{auxiliarytraining}
Linfeng Zhang, Muzhou Yu, Tong Chen, Zuoqiang Shi, Chenglong Bao, and Kaisheng
  Ma.
\newblock Auxiliary training: Towards accurate and robust models.
\newblock In {\em Proceedings of the IEEE/CVF Conference on Computer Vision and
  Pattern Recognition}, pages 372--381, 2020.

\bibitem{randomerasing}
Zhun Zhong, Liang Zheng, Guoliang Kang, Shaozi Li, and Yi~Yang.
\newblock Random erasing data augmentation.
\newblock In {\em Proceedings of the AAAI conference on artificial
  intelligence}, volume~34, pages 13001--13008, 2020.

\bibitem{INQ}
Aojun Zhou, Anbang Yao, Yiwen Guo, Lin Xu, and Yurong Chen.
\newblock Incremental network quantization: Towards lossless cnns with
  low-precision weights.
\newblock {\em arXiv preprint arXiv:1702.03044}, 2017.

\bibitem{vit_prune1}
Mingjian Zhu, Kai Han, Yehui Tang, and Yunhe Wang.
\newblock Visual transformer pruning.
\newblock {\em arXiv e-prints}, pages arXiv--2104, 2021.

\end{thebibliography}
}








\end{document}